\documentclass[10pt,twocolumn,letterpaper]{article}

\usepackage{wacv}              % To produce the CAMERA-READY version
\usepackage{graphicx}
\usepackage{amsmath}
\usepackage{amssymb}
\usepackage{booktabs}
\usepackage{xcolor}
\usepackage{multirow}
\usepackage[pagebackref,breaklinks,colorlinks]{hyperref}

\usepackage[capitalize]{cleveref}
\crefname{section}{Sec.}{Secs.}
\Crefname{section}{Section}{Sections}
\Crefname{table}{Table}{Tables}
\crefname{table}{Tab.}{Tabs.}

\begin{document}

%\title{SAM-Mamba: Mamba Guided SAM Architecture for efficient Out-of-Distribution Polyp Segmentation}
\title{SAM-Mamba: Mamba Guided SAM Architecture for Generalized Zero-Shot Polyp Segmentation}

\author{\parbox{16cm}{\centering
    {\large Tapas Kumar Dutta$^{1}$, Snehashis Majhi$^{2}$, Deepak Ranjan Nayak$^{3}$, and Debesh Jha$^4$}\\
    {\normalsize
    $^1$ University of Surrey, United Kingdom\quad
    $^2$ Côte d'Azur University, France \\\quad
    $^3$ Malaviya National Institute of Technology Jaipur, India \quad
    $^4$ University of South Dakota, USA}}\\
    \vspace{0.3cm}
    \small{\texttt{drnayak.cse@mnit.ac.in}}
    \vspace{-0.7cm}
}

\maketitle

%%%%%%%%% ABSTRACT
\begin{abstract}
Polyp segmentation in colonoscopy is crucial for detecting colorectal cancer. However, it is challenging due to variations in the structure, color, and size of polyps, as well as the lack of clear boundaries with surrounding tissues. Traditional segmentation models based on Convolutional Neural Networks (CNNs) struggle to capture detailed patterns and global context, limiting their performance. Vision Transformer (ViT)-based models address some of these issues but have difficulties in capturing local context and lack strong zero-shot generalization. To this end, we propose the Mamba-guided Segment Anything Model (SAM-Mamba\footnote[2]{Code, Models: \url{https://github.com/TapasKumarDutta1/SAM_Mamba_2025}}) for efficient polyp segmentation. Our approach introduces a Mamba-Prior module in the encoder to bridge the gap between the general pre-trained representation of SAM and polyp-relevant trivial clues. It injects salient cues of polyp images into the SAM image encoder as a domain prior while capturing global dependencies at various scales, leading to more accurate segmentation results.  Extensive experiments on five benchmark datasets show that SAM-Mamba outperforms traditional CNN, ViT, and Adapter-based models in both quantitative and qualitative measures. Additionally, SAM-Mamba demonstrates excellent adaptability to unseen datasets, making it highly suitable for real-time clinical use.

\end{abstract}

%%%%%%%%% BODY TEXT
\section{Introduction}
\label{sec:intro}
Colorectal Cancer (CRC) is considered the most prevalent gastrointestinal cancer and ranks as the third most common type of cancer across the globe. It often develops as a growth known as polyps in the colon's inner wall, leading to CRC if left undetected and untreated. Therefore, early detection and timely treatment are indispensable to preventing CRC, thereby reducing mortality rates. Colonoscopy is a widely adopted procedure for detecting and resecting polyps in the colon. However, the identification and segmentation of polyps through manual inspection of colonoscopy images is tedious and needs skilled and highly experienced clinicians. In addition, the tiny-sized polyps are likely to be overlooked during manual examination. Therefore, there is a strong need for the development of automated polyp segmentation methods to improve detection performance and potentially assist clinicians during the colonoscopy examination.

The last decade has witnessed a significant stride towards the development of medical image segmentation methods~\cite{chen2021transunet, gao2021utnet,lin2022ds, salpea2022medical, azad2024medical} using deep learning architectures, particularly encoder-decoder Convolutional Neural Networks (CNN) \cite{ronneberger2015u,zhou2018unet++}. Early polyp segmentation methods were based on the popular U-Net architecture \cite{ronneberger2015u} with some auxiliary components such as residual and dense connections and attention mechanisms, which include U-Net \cite{zhou2018unet++}, and ResUNet++ \cite{jha2019resunet++}. These methods lack the ability to deal with  
the crucial boundary information. To handle this issue, several segmentation methods such as FCN \cite{fang2019selective}, PraNet \cite{fan2020pranet},  CFA-Net \cite{zhou2023cross}, and MEGANet \cite{bui2024meganet} were designed. On the other hand, a few methods, including MSNet \cite{zhao2021automatic}, M$^2$UNet \cite{trinh2023m2unet}, and M$^2$SNet \cite{zhao2023m}, were introduced to deal with the scale diversity between various polyps. Although these methods achieved great success in segmenting polyps and their boundaries, they failed to draw global feature relationships which are crucial to detect complex and tiny polyps. Moreover, polyp segmentation still remains challenging due to the high similarity between polyps and the surrounding tissues in color and texture, significant shape and size variations among polyps, and indistinct boundaries. In addition, the repeated downscaling operations cause difficulties in recovering tiny polyps. Further, these models often fail to generalize better on unseen data captured through various image acquisition devices due to learning of varying image features.

Transformers have achieved incredible success in a wide range of computer vision tasks, including medical image analysis, because of their capability to model wide-range feature dependencies via self-attention \cite{dosovitskiy2021,shamshad2023transformers}. Following their success, efforts have been made towards the development of transformer-based segmentation methods such as TransUNet~\cite{chen2021transunet}, and UNETR~\cite{hatamizadeh2022unetr}. However, these methods limit their capability to capture local contextual information. Although a few recent works introduced convolutional layers in encoder and/or decoder to overcome the above issues, their application in the realm of polyp classification remains unexplored. Recently, a few transformer-based methods such as PVT-Cascade \cite{rahman2023medical} and CTNet \cite{xiao2024ctnet} were proposed and their performance was shown to be impressive on seen datasets. However, their generalization performance on unseen datasets is limited and the feature learning ability of such models is yet to be enhanced further to meet real-time clinical requirements.

Segment Anything Model (SAM) has recently been introduced as a foundational model for image segmentation and is well known for its impressive zero-shot generalization performance on unseen datasets \cite{kirillov2023segany}. However, SAM exhibits lower performance when directly applied to medical image segmentation, including polyp segmentation, due to the lack of domain-specific knowledge \cite{zhou2023can}. Meanwhile, the fine-tuning of SAM on medical data leads to high computational costs and memory requirements \cite{wu2023medical}. Adapter modules have emerged recently to overcome the above limitation and adapt to target tasks with less effort \cite{chen2022vision,wu2023medical}. More recently, Mamba leveraging State Space Models (SSM) has gained remarkable attention to effectively model long-range dependencies in sequential data with exceptional computational speed and memory efficiency \cite{zhu2024vision}. Some Mamba-based segmentation models include U-Mamba \cite{ma2024u} and SegMamba \cite{xing2024segmamba}. The generalization performance of these models still remains unexplored. Inspired by the recent success of these techniques, a Mamba-based prior coupled with SAM is proposed for effective polyp segmentation on both seen and unseen datasets. Specifically, a Mamba-Prior module, consisting of a Multi-scale Spatial Decomposition (MSD) and Mamba block, is designed to fully capture global contexts at various scales, thereby facilitating the segmentation of polyps of diverse scales and their complex boundaries. 
The extensive evaluations across five datasets demonstrate the effectiveness of the proposed SAM-Mamba model and its ability to achieve better zero-shot generalization performance compared to state-of-the-art CNN, ViT, and Adapter-based models. 

Our contributions are outlined as follows:
\begin{itemize}
\item We introduce a Mamba-based prior in SAM (SAM-Mamba) for enhanced generalized zero-shot polyp segmentation that leverages the learning capability of traditional SAM by effectively capturing multi-scale and global contextual cues of polyp image. \textit{To the best of our knowledge, this is the first attempt to explore the effectiveness of Adapter and Mamba within SAM for polyp segmentation.}
\item We propose a Mamba-Prior module comprising an MSD followed by Mamba blocks to inject the learned features into the SAM encoder. The former block aids in learning spatial features at various scales, while the latter block comprehensively captures broader contexts within feature maps, enriching learned feature representations and thereby segmenting complex polyps and their boundaries effectively.
\item We evaluate SAM-Mamba on five different benchmark datasets and compare its effectiveness against state-of-the-art polyp segmentation methods. Also, we perform ablative experiments to derive the significance of different components of the proposed module. The results demonstrate that the SAM-Mamba enjoys effective zero-shot generalization capabilities. 
\end{itemize}
%-------------------------------------------------------------------------
\begin{figure*}
\centering
\includegraphics[width=\linewidth]{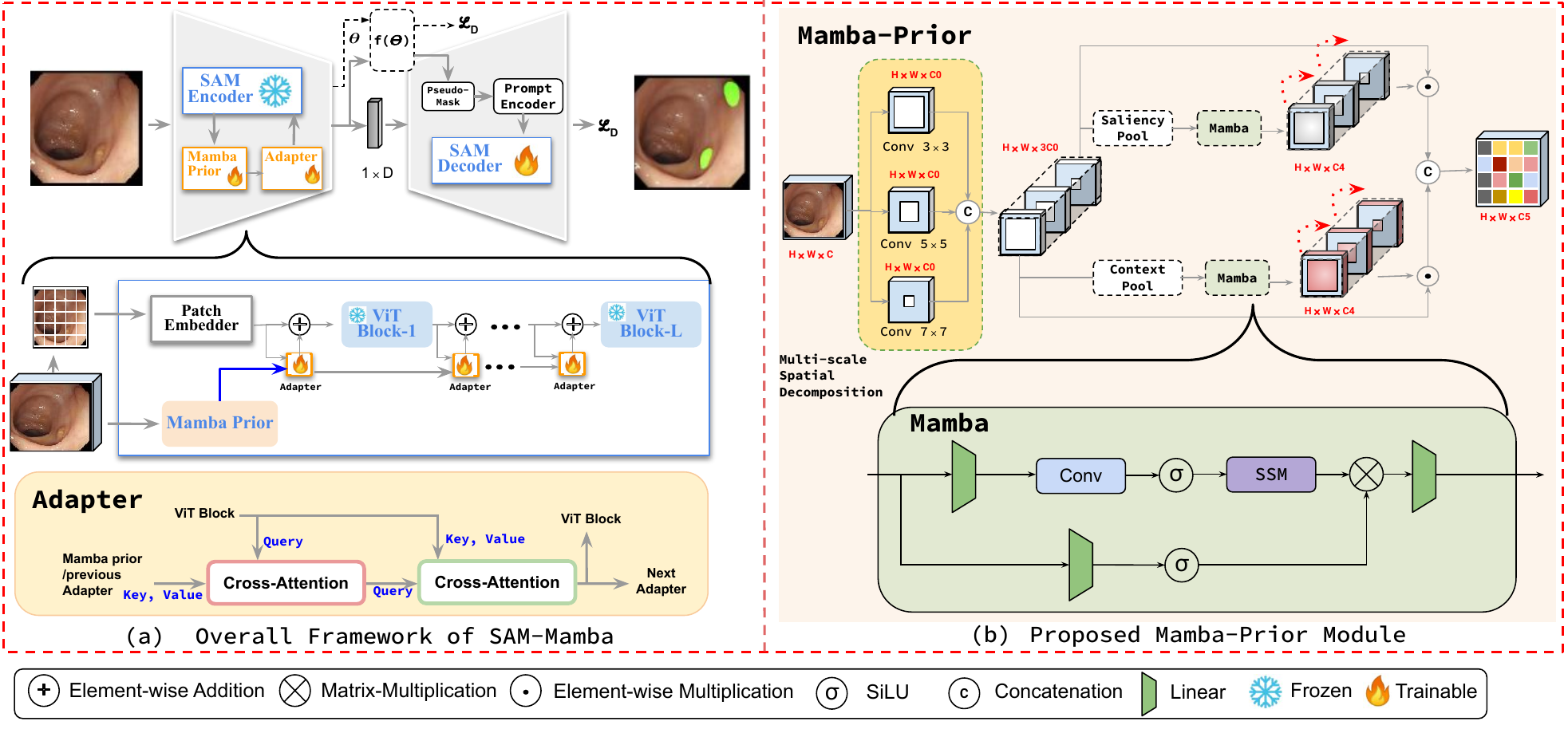}
% Copy of Tapas Mamba-SAM-2.pdf
\vspace{-0.4cm}
\caption{\small  \textbf{Overview of the SAM-Mamba framework} for polyp segmentation. The architecture constitutes the SAM backbone with the Mamba-Prior module and Adapter-based fine-tuning to enhance adaptability for polyp segmentation, addressing challenges like zero-shot feature transfer-ability, computational cost, and prompt dependency in SAM.}
\label{fig2}
\vspace{-0.4cm}
\end{figure*}

%-------------------------------------------------------------------------
% Table 1: Kvasir-SEG (seen)
\section{Related Work}
The initial efforts in the realm of polyp segmentation have been made with the most popular encoder-decoder-based CNN architecture, U-Net. For instance, Zhou et al. \cite{zhou2018unet++} developed UNet++ with a series of nested dense connections between encoder and decoder sub-networks for polyp segmentation. Jha et al. \cite{jha2019resunet++} proposed an improved ResUNet model known as ResUNet++ by introducing additional attention blocks and pooling layers for accurate segmentation of colorectal polyps. To establish the polyp area-boundary relationships, Fang et al. \cite{fang2019selective} proposed a Selective Feature Aggregation (SFA) network that uses a convolutional-based shared encoder, dual decoders, and a boundary-sensitive loss function. Ping et al. \cite{fan2020pranet} proposed PraNet that employs parallel partial decoders and reverse attention modules to refine segmentation boundaries and enhance polyp segmentation accuracy. Wei et al. \cite{wei2021shallow} devised SANet based on color exchange operation, shallow attention module, and probability correction strategy to improve polyp segmentation accuracy and address color inconsistency, small polyp degradation, and pixel imbalance. In \cite{zhao2021automatic},  a Multi-scale Subtraction Network (MSNet) was designed by concatenating multiple subtraction units pyramidally to address the scale diversity of polyps and introducing a loss function for detailed-to-structure supervision. An improved version of MSNet known as M$^2$SNet was proposed in \cite{zhao2023m} that leverages intra-layer multi-scale subtraction unit along with inter-layer multi-scale subtraction unit for efficient polyp segmentation. Mau et al. \cite{nguyen2023pefnet} proposed to utilize an EfficientNetV2 backbone-based U-Net with a new positional embedding feature block to 
enhance feature transfer and improve polyp segmentation accuracy and generalization. Zhou et al.  \cite{zhou2023cross} introduced a Cross-level Feature Aggregation Network  (CFA-Net) that integrates boundary-aware features and cross-level feature fusion to address scale variation and boundary ambiguity in polyps. Recently, a Multi-scale Edge-Guided Attention Network called MEGANet was designed in \cite{bui2024meganet} that integrates edge-guided attention modules between encoder and decoder to retain edge information, improving the segmentation polyps with weak boundaries.

Transformers have shown their prominence in medical image segmentation due to their potential to draw global contextual details and, therefore, have recently been exploited in polyp segmentation tasks. Trinh et al. \cite{trinh2023m2unet} devised M$^2$Unet, which uses a hybrid CNN-Transformer encoder and integrates multi-scale upsampling block to combine multi-level decoder information, enhancing local and global feature representation of polyps. To ensure robust feature learning ability for polyp segmentation, Rahman et al. \cite{rahman2023medical} designed PVT-Cascade using a hierarchical cascaded attention-based decoder, which integrates multi-scale features with attention gates and convolutional attention modules, thereby enhancing both global and local contexts. In a recent contribution, a Contrastive Transformer Network (CTNet) \cite{xiao2024ctnet} was designed with different modules such as a contrastive transformer backbone, self-multiscale interaction module, and collection information modules to obtain stable polyp segmentation results and better generalization performance. Despite the significant progress, these methods still struggle with the challenges that lie in the polyps and their camouflage properties. Additionally, there exists abundant room for improving the generalization abilities.  
\section{Methodology}
In this section, we introduce our SAM-Mamba framework step-by-step for polyp segmentation. Since our SAM-Mamba generously extends the notion of SAM backbone, we first elaborate on the preliminaries of SAM, which emerged as a general-purpose object segmentation model. Then, we discuss our SAM-Mamba, which proliferates the adaptability of SAM in polyp segmentation via a novel Mamba-Prior module, prompting to handle the critical challenges of the task.

\subsection{Preliminaries of SAM}
The SAM architecture consists of three primary components: an image encoder, a prompt encoder, and a mask decoder. \textbf{Image Encoder:} The image encoder is built on a standard ViT architecture that has been pre-trained using Masked Autoencoders (MAE). Specifically, it utilizes the \texttt{ViT-H/16} variant, which incorporates $14 \times 14$ windowed attention along with four equally spaced global attention blocks. The encoder outputs a $16 \times $ down-sampled embedding of the input image. \textbf{Prompt Encoder:} The prompt encoder can handle either sparse (e.g., points, boxes) or dense (e.g., masks) prompts. Here, the sparse encoder encodes points and boxes using positional encoding combined with learned embedding specific to each prompt type. \textbf{Mask Decoder:} The mask decoder is a modified Transformer decoder block including a dynamic mask prediction head, which employs two-way cross-attention mechanisms, facilitating the learning of interactions among the prompt and image embeddings. After processing, SAM up samples the image embeddings, and the output tokens are mapped by MLP to a dynamic linear classifier to predict the target mask for the given image. Thanks to these three components of SAM as they help to achieve promising results on general object segmentation tasks, but there exist several challenges while adapting SAM to the polyp segmentation task.

First, \textbf{inferior transfer learning ability:} by following the traditional full fine-tuning strategy, SAM may result in overfitting, forgetting, or even feature degradation, especially for large pre-trained models when the downstream datasets are not sufficiently large and diverse. Second, \textbf{increased computational cost:} SAM processes $4\times $ higher resolution input images compared to classical ViT, which increases the number of patches, thereby increasing the computation cost of full fine-tuning SAM with a factor of $4$. Further, \textbf{dependency of SAM in \texttt{point, box, text} prompts:} SAM requires a prompt or a set of prompts to produce a segmentation mask; however, in the case of most polyp segmentation models the input is simply a polyp image and the output is a segmentation mask. Thus, utilizing SAM in its original form for the polyp segmentation task is still far from feasible.
\subsection{SAM-Mamba}
Our objective in SAM-Mamba shown in Figure~\ref{fig2}{\color{red} a} is to enhance the adaptability of the SAM architecture for polyp segmentation tasks through effective and lightweight fine-tuning. Unlike traditional full fine-tuning methods that optimize all parameters, we maintain the pre-trained SAM parameters frozen and follow an Adapter-based fine-tuning. The Adapter serves as a bottleneck model and consists of low-level parameters to adapt the polyp image domain. However, in contrast to general images, the polyp image segmentation task entails trivial attention to several key attributes, \textit{i.e.} to distinguish between polyp region and neighborhood pixel in terms of color, shape, and indistinguishable boundaries. Thus, a miniaturized adapter with limited learnable parameters may not be adequate to learn the critical discriminative feature representations for the polyp segmentation task. 
% It need dedicated method.

\vspace{-0.5cm}
% Thus, to bridge the gap between the general pretrained reprentation of SAM and 
\subsubsection{Mamba-Prior Module}
\vspace{-0.1cm}
In order to bridge the gap between the general pre-trained representation of SAM and polyp-relevant trivial clues, a Mamba-Prior module outlined in Figure~\ref{fig2}{\color{red} b} is proposed. For injecting salient cues of polyp images into SAM image encoder as a domain prior, it adopts three strategies, \textbf{(i) Multi-scale Spatial Decomposition:} encodes the low-level features (such as size, shape, boundary) of polyp region at various spatial scales and thereby, helping to analyze spatial fine-grained to coarse semantics, \textbf{(ii) Channel Saliency and Context Accumulation:} for each scale of spatial decomposition, it extracts the salient and contextual cues in a mutually exclusive manner and accumulates them along the channel depth, \textbf{(iii) Mamba Channel Interaction:} leverages Mamba layer to capture the long-range inner variations in salient and contextual channels encoded at multiple spatial scales. The Mamba-Prior with these functional strategies is learned alongside the adapters to inject polyp domain-specific critical cues to the SAM encoder. The functional details of each strategy are as follows:
\paragraph{Multi-scale Spatial Decomposition (MSD):} It decomposes the input polyp image $I \in \mathbb{R}^{H \times W \times C}$ to multiple spatial scales by processing it in parallel convolution layers with different receptive fields ($k$), where $k\in \{ 3, 5, 7\}$. The resultant maps of an arbitrary convolution layer can be denoted as $M_k$, $\exists k \in \{ 3, 5, 7\}$. It is to be noted that the resultant maps  $M_3, \ldots, M_7 \in \mathbb{R}^{H \times W \times C0}$ are \texttt{padded} to match the original image size with a different number of filters ($C0$). In order to build a spatial multi-scale feature pyramid $M^* \in \mathbb{R}^{H \times W \times 3\dot C0}$, the resultant maps are stacked via \texttt{concatenate operator} along channel depths and we ensure the coarse-grained map $M_7$ lies in the top and fine-grained map $M_3$ remains in the bottom of the pyramid. 
Consequently, this enables the model to analyze the polyp region in an orderly decomposition. %The $M^*$ is mathematically expressed as 
% \begin{equation}
%     M^*=Concat(M_3,M_5,M_7)
% \end{equation}
% where,
% \begin{equation}
%     M_k=Conv_{k\times k}(I).
% \end{equation}
\paragraph{Channel Saliency and Context Accumulation :} It focuses on extracting and accumulating salient and broader contextual features within $M^* \in \mathbb{R}^{H \times W \times 3\dot C0}$. For the saliency and contextual extraction, we apply standard \texttt{global-max-pool} and \texttt{global-average-pool} operators that generates $M^S \in \mathbb{R}^{1 \times 3\dot C0}$ and $M^C \in \mathbb{R}^{1 \times 3\dot C0}$, respectively. 
% In both the pool operation the kernel size  is maintained to $2 \times 2$ with stride $2$.
\paragraph{Mamba Channel Interaction :} Mamba demonstrates a robust ability to manage long sequence data with linear computational complexity, thereby we leverage its effectiveness for capturing intra-pixel interactions of salient and contextual maps \textit{i.e.,} $M^S$ and $M^C$ with two parallel Mamba layers. Primarily, Mamba is allowed to independently encode the dependency between the multi-scale channel distribution of $M^S$ and $M^C$. Mamba utilizes a gated mechanism, as shown in Figure~\ref{fig2}, to further refine feature representations. For the given input  $M^S$ and $M^C$ to Mamba, the resultant feature maps can be obtained as follows:
\begin{equation}\label{eq:1}
%\hat{M}^S &= \phi(M^S) , \  \hat{M}^C = \phi(M^C)\\
M^S_o = \phi\big(\text{SSM}(\sigma(\text{Conv}(\phi(M^S)))) \otimes \sigma(\phi(M^S))\big)
\end{equation}
%\hat{M^C} &= \text{Linear}\big(\text{SSM}(\text{SiLU}(\text{Conv}(\hat{M^C} \times \text{Linear}(\text{SiLU}(\hat{M^C})))))\big)
\begin{equation}\label{eq:2}
M^C_o = \phi\big(\text{SSM}(\sigma(\text{Conv}(\phi(M^C)))) \otimes \sigma(\phi(M^C))\big)
\end{equation}

where, $\phi(.)$ indicates a \texttt{linear} layer, $\sigma$ indicates \texttt{SiLU} activation and $\otimes$ denotes the matrix multiplication.
The dense and orderly intra-channel interaction encoding in $M^S_o$ and $M^C_o$ enables understanding critical cues of polyp regions in multi-scale pooled feature map $M^S$ and $M^C$. However, through Mamba gating, there is a possibility of forgetting fine-grained and sparse cues. Thus, we have multiplied the original multi-scale feature map of the polyp image with the resultant of Mamba $M^S_o$ and $M^C_o$ through a \texttt{skip-connection} and thereafter the multiplied resultants are concatenated across channel depth for obtaining domain prior embedded feature map $M^D \in \mathbb{R}^{H \times W \times C5}$, which is formally expressed as
\begin{equation}
    M^D=\texttt{Concat}(M^S_o \odot M^*, M^C_o \odot M^*)
\end{equation}

\vspace{-0.5cm}
\subsubsection{Adapter}
\vspace{-0.1cm}
The adapter module incorporated in our framework draws inspiration from~\cite{chen2022vision} and is mainly based
on two sequential \texttt{cross-attention}, as shown in Figure~\ref{fig2}{\color{red} a}. The first and second \texttt{cross-attention} are used to enhance the multi-scale features and then to inject the Mamba-Prior into ViT blocks, respectively. This injection ensures that the feature distribution of the ViT block will not be modified drastically, thus making better use of the pre-trained ViT.

% The adapter inspired by \cite{chen2022vision} is implemented in 2 distinct stages:
% \textbf{Stage 1: } The intermediate features maps from a ViT block is extracted and used as a \textbf{Query} and the features from mamba prior for the first adapter and the output of previous adapter is used as \textbf{Key} and \textbf{Value} to find features not that can complement the ViT extracted features. For the first adapter to inject information extracted from mamba prior we use a cross-attention

% \textbf{Stage 2: }With the complementing features extracted the question now becomes how to integrate the features with the ViT extracted feature maps. To this end we employ a cross-attention with output of previous cross-attention as \textbf{Query} and the feature maps from ViT block as \textbf{Key} and \textbf{Value}. 

\vspace{-0.3cm}
\subsubsection{SAM Decoder}
\vspace{-0.1cm}
For the mask decoder we adopt the architecture proposed by Kirillov \textit{et al.} \cite{kirillov2023segany} that utilizes prompts such as bounding box, masks, point, or text to further enrich the encoder extracted features for segmentation. However, this prevents the model from being used without these prompts. To this end, our \textbf{SAM-Mamba} first extracts a pseudo mask from $f(\theta)$ which inputs the features extracted by the SAM encoder by training the \textbf{Mamba-Prior} and \textbf{Adapter}, optimizes the model with $L_D$. Subsequently, the pseudo mask obtained from $f(\theta)$ is fed to the decoder as a prompt to refine the mask further by training the \textbf{Mamba-Prior}, \textbf{Adapter} and \textbf{SAM-Decoder} using $L_D$ for supervision.

\vspace{-0.5cm}
\subsubsection{Objective Function}
\vspace{-0.1cm}
Our proposed SAM-Mamba along with its functional blocks \textit{i.e.} Mamba-prior, adapter, $f(\theta)$, and SAM decoder is jointly trainable with a loss function is defined as a combination of Dice loss and weighted Binary Cross Entropy (BCE) loss, ${L}_D=L_w^{\text{Dice}}+L_w^{\text{BCE}}$. The Dice loss enhances the importance of hard pixels by increasing their weights, whereas the BCE loss places more emphasis on hard pixels instead of treating all pixels equally. Since  SAM decoder, heavily relies on the pseudo mask generated by $f(\theta)$, we follow a two-stage training regime, wherein the \textbf{Stage 1} the $f(\theta)$ is allowed to train for some iterations and thereafter training of SAM-decoder in \textbf{Stage 2} is initiated. The detailed training regime is described as in below,

\textbf{Stage 1:} The adapters within the image encoder are trained with deep supervision using a secondary output ($S^{\text{up}}_{Encoder}$) directly from the encoder as indicated by broken lines in Figure \ref{fig2}. The loss function for this stage is given by:
\begin{equation}
    L_{\text{stage1}} = L_D(G, S^{\text{up}}_{Encoder})
\end{equation}
where \( S^{\text{up}}_{Encoder} \) is the up-sampled side-output from the image encoder and supervised with the ground-truth $G$.

\textbf{Stage 2:} In the subsequent stage, the entire model, i.e. both the mask decoder and the image encoder's adapter, is trained with full supervision, as denoted by solid lines in Figure \ref{fig2}. The total loss in this stage is calculated as:
\begin{equation}
    L_{\text{stage2}} = L_D(G, S_{Decoder})
\end{equation}
Here, \( S_{Decoder} \) and \(S^{\text{up}}_{Encoder} \) are the mask decoder and the up-sampled outputs from the image encoder, respectively, and are compared to the ground-truth.
%-----------------------------------------------------
\section{Experiments}
\subsection{Datasets and Evaluation Metrics}
\paragraph{Datasets :} To evaluate the performance of SAM-Mamba we conduct experiments on five challenging polyp segmentation datasets: ETIS~\cite{silva2014toward}, CVC-ColonDB~\cite{tajbakhsh2015automated}, EndoScene~\cite{vazquez2017benchmark}, Kvasir-SEG~\cite{jha2020kvasir}, and CVC-ClinicDB~\cite{bernal2015wm}. 
%The first four are standard benchmarks, and the last one is the largest-scale challenging dataset, recently released. 
To ensure a fair comparison and exhibit zero-shot generalization capabilities, we adopt the same experimental setup as in PraNet\cite{fan2020pranet}. In the specified setting, 1,450 images are selected for the training set, of which 900 images are collected from Kvasir-SEG, and 550 images are collected from the CVC-ClinicDB dataset. The remaining 100 images from Kvasir-SEG and 62 images from CVC-ClinicDB are kept for the testing set. In addition, we adopt 380 images from CVC-ColonDB, 196 images from ETIS, and 60 images from the CVC-300 (test set of EndoScene) for testing. This configuration poses various challenges due to varying resolutions across different datasets and the varied image acquisition devices.
\paragraph{Evaluation Metrics :}
To perform a thorough evaluation and comparison, we adopt six different metrics: Dice, IoU, S-measure ($S_{\alpha}$) \cite{fan2017structure}, F-measure ($F_{\beta}^{w}$) \cite{margolin2014evaluate}, E-measure ($E_{\phi}^{\text{max}}$) \cite{fan2018enhanced}, and Mean Absolute Error (MAE) adhering to established state-of-the-art (SOTA) approaches. It is worth noting here that the mean of Dice and IoU is denoted as mDice and mIoU in our study. The details of these metrics are provided in \cite{fan2020pranet,zhao2021automatic}.

\begin{table*}[h!]
\centering
\caption{Quantitative results comparison of SAM-Mamba with SOTA methods on Kvasir-SEG and CVC-ClinicDB datasets (seen). `↑' and `↓ ' represent that larger or smaller scores are better. `\textcolor{red}{Red}' and `\textcolor{blue}{Blue}' color fonts indicate the best and second best scores.}
\label{tab:combined_performance}
\resizebox{\textwidth}{!}{%
\begin{tabular}{|l|c|c|c|c|c|c|c|c|c|c|c|c|}
\hline
\multirow{2}{*}{Methods} & \multicolumn{6}{c|}{Kvasir-SEG (Seen)} & \multicolumn{6}{c|}{CVC-ClinicDB (Seen)} \\ \cline{2-13} 
 & mDice ↑ & mIoU ↑ & $F_{\beta}^{w}$ ↑ & $S_{\alpha}$ ↑ & $E_{\phi}^{\text{max}}$ ↑ & MAE ↓ & mDice ↑ & mIoU ↑ & $F_{\beta}^{w}$ ↑ & $S_{\alpha}$ ↑ & $E_{\phi}^{\text{max}}$ ↑ & MAE ↓ \\ \hline
U-Net \cite{ronneberger2015u}       & 81.8 & 74.6 & 79.4 & 85.8 & 89.3 & 5.5 & 82.3 & 75.5 & 81.1 & 88.9 & 95.4 & 1.9 \\
U-Net++ \cite{zhou2018unet++}       & 82.1 & 74.3 & 80.8 & 86.2 & 91.0 & 4.8 & 79.4 & 72.9 & 78.5 & 87.3 & 93.1 & 2.2 \\ 
SFA \cite{fang2019selective}        & 72.3 & 61.1 & 67.0 & 78.2 & 84.9 & 7.5 & 70.0 & 60.7 & 64.7 & 79.3 & 88.5 & 4.2 \\ 
PraNet \cite{fan2020pranet}         & 89.8 & 84.0 & 88.5 & 91.5 & 94.8 & 3.0 & 89.9 & 84.9 & 89.6 & 93.6 & 97.9 & 0.9 \\ 
SANet \cite{wei2021shallow}         & 90.4 & 84.7 & 89.2 & 91.5 & 95.3 & 2.8 & 91.6 & 85.9 & 90.9 & 93.9 & 97.6 & 1.2 \\ 
MSNet \cite{zhao2021automatic}      & 90.7 & 86.2 & 89.3 & 92.2 & 94.4 & 2.8 & 92.1 & 87.9 & 91.4 & 94.1 & 97.2 & 0.8 \\ 

Polyp-PVT\cite{dong2021polyp} & \textcolor{blue}{91.7} & \textcolor{blue}{86.4} & \textcolor{blue}{91.1} & 92.5 & 95.6  &  \textcolor{red}{2.3} & 93.7 & \textcolor{blue}{88.9} & 93.6 & 94.9 & 98.5  & \textcolor{red}{0.6}\\

PEFNet \cite{nguyen2023pefnet}      & 89.2 & 83.3 & –    & –    & –    & 2.9 & 86.6 & 81.4 & –    & –    & –    & 1.0 \\ 
M$^2$UNet \cite{trinh2023m2unet}    & 90.7 & 85.5 & –    & –    & –    & \textcolor{blue}{2.5} & 90.1 & 85.3 & –    & –    & –    & 0.8 \\ 
M$^2$SNet \cite{zhao2023m}          & 91.2 & 86.1 & 90.1 & 92.2 & 95.3 & \textcolor{blue}{2.5} & 92.2 & 88.0 & 91.7 & 94.2 & 97.0 & 0.9 \\ 
PVT-Cascade \cite{rahman2023medical} & 91.1 & 86.3 & 90.6 & 91.9 & \textcolor{blue}{96.1} & \textcolor{blue}{2.5} & 91.9 & 87.2 & 91.8 & 93.6 & 96.9 & 1.3 \\ 
CTNet \cite{xiao2024ctnet}          & \textcolor{blue}{91.7} & 86.3 & 91.0 & \textcolor{blue}{92.8} & 95.9 & \textcolor{red}{2.3} & 93.6 & 88.7 & 93.4 & \textcolor{blue}{95.2} & 98.3 & \textcolor{red}{0.6} \\ 
CFA-Net \cite{zhou2023cross}        & 91.5 & 86.1 & 90.3 & 92.4 & \textcolor{red}{96.2} & \textcolor{red}{2.3} & 93.3 & 88.3 & 92.4 & 95.0 & \textcolor{red}{98.9} & \textcolor{blue}{0.7} \\ 
MEGANet \cite{bui2024meganet}       & 91.3 & 86.3 & 90.7 & 91.8 & 95.9 & \textcolor{blue}{2.5} & \textcolor{blue}{93.8} & \textcolor{red}{89.4} & \textcolor{blue}{94.0} & 95.0 & \textcolor{blue}{98.6} & \textcolor{red}{0.6} \\\hline

%\textbf{SAM-Mamba End-to-end}       & \textcolor{red}{91.8} & \textcolor{red}{87.0} & \textcolor{red}{94.2} & \textcolor{red}{93.0} & \textcolor{blue}{95.9} & \textcolor{blue}{2.5} & \textcolor{red}{93.8} & \textcolor{blue}{88.5} & \textcolor{red}{94.2} & \textcolor{red}{95.3} & 97.4 & \textcolor{red}{0.7} \\ 
\textbf{SAM-Mamba}         & \textcolor{red}{92.4} & \textcolor{red}{87.3} & \textcolor{red}{94.2} & \textcolor{red}{93.6} & \textcolor{blue}{96.1} & \textcolor{blue}{2.5} & \textcolor{red}{94.2} & 88.7 & \textcolor{red}{94.3} & \textcolor{red}{95.5} & 98.2 & \textcolor{red}{0.6} \\ \hline
\end{tabular}%
}
\vspace{-0.5cm}
\end{table*}

\subsection{Implementation Details}
The SAM-Mamba model is implemented using PyTorch and accelerated with an NVIDIA A100 GPU. All inputs are resized to \( 352 \times 352 \) pixels. A multi-scale training strategy with scales \{0.75, 1, 1.25\} is used for data augmentation. Adam optimizer is employed with a learning rate of \( 1 \times 10^{-5} \) to train the model. The model is trained upto 200 epochs.

\subsection{Quantitative Comparison}
%In this section, we evaluate the performance of the SAM-Adapter model, focusing on its learning ability, generalization capability, complexity, and qualitative results. These evaluations are crucial in determining the effectiveness of SAM-Adapter in comparison to other state-of-the-art (SOTA) models. 
To verify the robustness of our SAM-Mamba, we extensively compare it with 14 SOTA segmentation methods, such as  U-Net\cite{ronneberger2015u} and U-Net++\cite{zhou2018unet++},  SFA\cite{fang2019selective},  PraNet\cite{fan2020pranet}, 
SANet\cite{wei2021shallow}, MSNet\cite{zhao2021automatic}, Polyp-PVT \cite{dong2021polyp}, PEFNet\cite{nguyen2023pefnet}, M$^2$UNet \cite{trinh2023m2unet}, PVT-Cascade \cite{rahman2023medical}, M$^2$SNet \cite{zhao2023m}, CFA-Net \cite{zhou2023cross}, CTNet \cite{xiao2024ctnet}, and MEGANet \cite{bui2024meganet}

We validate the learning ability of SAM-Mamba on two benchmark datasets, Kvasir-SEG and CVC-ClinicDB. As detailed in Table~\ref{tab:combined_performance}, SAM-Mamba is rigorously compared against SOTA CNN and ViT-based segmentation models. On the challenging Kvasir-SEG dataset, SAM-Mamba surpasses the competitive CTNet~\cite{xiao2024ctnet}, achieving an impressive 92.4\% in the \textbf{mDice} metric. Similarly, it demonstrates exceptional and consistent performance on CVC-ClinicDB, outperforming peers in key metrics such as \textbf{mIoU}, \textbf{$F_{\beta}^{w}$}, and \textbf{$S_{\alpha}$}, underscoring its robustness and superiority in polyp segmentation. However, in terms of \textbf{$E_{\phi}^{\text{max}}$}, scores of 96.1\% and 95.5\% highlight slight room for improvement in edge detection. Its \textbf{MAE} values of 2.5 and 0.6, while competitive, suggest potential for further refinement. Overall, SAM-Mamba excels across most metrics, solidifying its dominance while leaving scope for advancements in edge sensitivity and error minimization. The model's learning trajectory is visually depicted in Figure~\ref{fig:heatmap}, showcasing intermediate feature maps, encoder outputs, decoder heatmaps, and refined segmentation masks alongside input and ground truth for comprehensive comparison. Notably, PEFNet results are sourced from M$^2$UNet, while other baselines are directly derived from their original works.

%-------------------------------------------------
\paragraph{Zero-shot Generalization Ability Verification:} 

A pivotal aspect of model evaluation lies in its ability to generalize effectively to unseen data in zero-shot scenarios, a critical requirement for real-world medical image segmentation applications. To assess this, we benchmark SAM-Mamba on three datasets: CVC-300, CVC-ColonDB, and ETIS, specifically testing its zero-shot generalization capabilities. As illustrated in Tables~\ref{tab:combined0} and \ref{tab:generalization2}, SAM-Mamba achieves remarkable performance, outperforming all SOTA models on the CVC-ColonDB and ETIS datasets by substantial margins of $+4\%$ and $+3.8\%$, respectively, in terms of \textbf{mIoU} metric. Comparable performance gains are observed across other metrics on these datasets, further validating the robustness and adaptability  of the proposed model in zero-shot generalization. %These results underscore the model's potential to bridge the gap between laboratory benchmarks and real-world clinical applications, making it a standout choice for unseen data scenarios in medical imaging.

%---------------------------------------------------

%-----------------------------------------------%

%-----------------------------------------------------%

%------------------------------------------------------%
\begin{table*}[h!]
\centering
\caption{Quantitative results comparison of SAM-Mamba with SOTA methods on CVC-300 and CVC-ColonDB datasets (unseen). `↑' and `↓ ' represent that larger or smaller scores are better. `\textcolor{red}{Red}' and `\textcolor{blue}{Blue}' color fonts indicate the best and second best scores.}
\label{tab:combined0}
\resizebox{\textwidth}{!}{%
\begin{tabular}{|l|c|c|c|c|c|c|c|c|c|c|c|c|}
\hline
\multirow{2}{*}{Methods} & \multicolumn{6}{c|}{CVC-300 (Unseen)} & \multicolumn{6}{c|}{CVC-ColonDB (Unseen)} \\ \cline{2-13} 
 & mDice ↑ & mIoU ↑ & $F_{\beta}^{w}$ ↑ & $S_{\alpha}$ ↑ & $E_{\phi}^{\text{max}}$ ↑ & MAE ↓ & mDice ↑ & mIoU ↑ & $F_{\beta}^{w}$ ↑ & $S_{\alpha}$ ↑ & $E_{\phi}^{\text{max}}$ ↑ & MAE ↓ \\ 
 \hline
U-Net\cite{ronneberger2015u} & 71.0 & 62.7 & 68.4 & 84.3 & 87.6 & 2.2 & 51.2 & 44.4 & 49.8 & 71.2 & 77.6 & 6.1 \\ 
U-Net++\cite{zhou2018unet++} & 70.7 & 62.4 & 68.7 & 83.9 & 89.8 & 1.8 & 48.3 & 41.0 & 46.7 & 69.1 & 76.0 & 6.4 \\ 
SFA\cite{fang2019selective} & 46.7 & 32.9 & 34.1 & 64.0 & 81.7 & 6.5 & 46.9 & 34.7 & 37.9 & 63.4 & 76.5 & 9.4 \\ 
PraNet\cite{fan2020pranet} & 87.1 & 79.7 & 84.3 & 92.5 & 97.2 & 1.0 & 70.9 & 64.0 & 69.6 & 81.9 & 86.9 & 4.5 \\ 
SANet\cite{wei2021shallow} & 88.8 & 81.5 & 85.9 & 92.8 & 97.2 & 0.8 & 75.3 & 67.0 & 72.6 & 83.7 & 87.8 & 4.3 \\ 
MSNet\cite{zhao2021automatic} & 86.9 & 80.7 & 84.9 & 92.5 & 94.3 & 1.0 & 75.5 & 67.8 & 73.7 & 83.6 & 88.3 & 4.1 \\ 
Polyp-PVT\cite{dong2021polyp} & 90.0 & 83.3 & 88.4 & 93.5 & 97.3 & \textcolor{blue}{0.7} & 80.8 & 72.7 & 79.5 & 86.5 & 91.3 & 3.1 \\
PEFNet\cite{nguyen2023pefnet} & 87.1 & 79.7 & – & – & – & 1.0 & 71.0 & 63.8 & – & – & – & 3.6 \\ 
M$^2$UNet\cite{trinh2023m2unet} & 89.0 & 81.9 & – & – & – & 0.7 & 76.7 & 68.4 & – & – & – & 3.6 \\ 
M$^2$SNet\cite{zhao2023m} & 90.3 & 84.2 & 88.1 & 93.9 & 96.5 & 0.9 & 75.8 & 68.5 & 73.7 & 84.2 & 86.9 & 3.8 \\ 
PVT-Cascade\cite{rahman2023medical} & 89.2 & 82.4 & 87.3 & 93.2 & 95.9 & 0.9 & 78.1 & 71.0 & 77.9 & 85.5 & 89.6 & 3.1 \\
CTNet\cite{xiao2024ctnet} & \textcolor{blue}{90.8} & \textcolor{blue}{84.4} & \textcolor{blue}{89.4} & \textcolor{red}{97.5} & 97.5 & \textcolor{red}{0.6} & \textcolor{blue}{81.3} & \textcolor{blue}{73.4} & \textcolor{blue}{80.1} & \textcolor{blue}{87.4} & \textcolor{blue}{91.5} & \textcolor{blue}{2.7} \\ 

CFA-Net\cite{zhou2023cross} & 89.3 & 82.7 & \textcolor{red}{93.8} & 87.5 & \textcolor{blue}{97.8} & 0.8 & 74.3 & 66.5 & 72.8 & 83.5 & 89.8 & 3.9 \\ 
MEGANet\cite{bui2024meganet} & 89.9 & 83.4 & 88.2 & 93.5 & 96.9 & \textcolor{blue}{0.7} & 79.3 & 71.4 & 77.9 & 85.4 & 89.5 & 4.0 \\\hline
%\textbf{SAM-Mamba End-to-end} & \textcolor{red}{91.1} & \textcolor{red}{85.8} & 88.6 & \textcolor{blue}{94.4} & \textcolor{red}{98.1} & \textcolor{red}{0.6} & \textcolor{red}{82.1} & \textcolor{red}{73.5} & \textcolor{red}{81.2} & \textcolor{red}{87.9} & \textcolor{red}{92.1} & \textcolor{red}{2.0} \\ \hline

\textbf{SAM-Mamba} & \textcolor{red}{92.0} & \textcolor{red}{86.1} &  88.8& \textcolor{blue}{94.6}& \textcolor{red}{98.1}& \textcolor{red}{0.6} & \textcolor{red}{85.3} & \textcolor{red}{77.1} & \textcolor{red}{85.6} &  \textcolor{red}{89.8} & \textcolor{red}{93.3} & \textcolor{red}{1.7}\\ \hline
\end{tabular}}
\end{table*}

%-----------------------------------------------------%
% \vspace{-0.5cm}
\begin{table}[h!]
\centering
\caption{Quantitative results comparison of SAM-Mamba with SOTA methods on ETIS dataset (unseen).} %`↑' and `↓ ' represent that larger or smaller scores are better. `\textcolor{red}{Red}' and `\textcolor{blue}{Blue}' color fonts indicate the best and second best scores. .}
\label{tab:generalization2}
\resizebox{\columnwidth}{!}{\begin{tabular}{|l|c|c|c|c|c|c|}
\hline
Methods & mDice ↑ & mIoU ↑ & $F_{\beta}^{w}$ ↑ & $S_{\alpha}$ ↑ & $E_{\phi}^{\text{max}}$ ↑ & MAE ↓ \\ \hline
U-Net\cite{ronneberger2015u}         & 39.8 & 33.5 & 36.6 & 68.4 & 74.0 & 3.6 \\
U-Net++\cite{zhou2018unet++}       & 40.1 & 34.4 & 39.0 & 68.3 & 77.6 & 3.5 \\ 
SFA\cite{fang2019selective}                 & 29.7 & 21.7 & 23.1 & 55.7 & 63.3 & 10.9 \\ 
PraNet \cite{fan2020pranet}        & 62.8 & 56.7 & 60.0 & 79.4 & 84.1 & 3.1 \\ 
SANet\cite{wei2021shallow}           & 75.0 & 65.4 & 68.5 & 84.9 & 89.7 & 1.5 \\ 
MSNet\cite{zhao2021automatic}     & 71.9 & 66.4 & 67.8 & 84.0 & 83.0 & 2.0 \\ 
Polyp-PVT \cite{dong2021polyp} &78.7 & 70.6 & 75.0 & 87.1 & 90.6  & 1.3\\
PEFNet   \cite{nguyen2023pefnet}      & 63.6 & 57.2 & –    & –    & –    & 1.9 \\ 
M$^2$UNet \cite{trinh2023m2unet}        & 67.0 & 59.5 & –    & –    & –    & 2.4 \\ 
M$^2$SNet \cite{zhao2023m} & 74.9 & 67.8 & 71.2 & 84.6 & 87.2 & 1.7 \\
PVT-Cascade \cite{rahman2023medical} & 78.6 & 71.2& 75.9 & 87.2& 89.6& \textcolor{blue}{1.3}\\
CTNet \cite{xiao2024ctnet} & \textcolor{blue}{81.0}& \textcolor{blue}{73.4}& \textcolor{blue}{77.6}& \textcolor{blue}{88.6} & \textcolor{blue}{91.3}& 1.4\\
CFA-Net \cite{zhou2023cross} & 73.2 & 65.5 & 69.3 & 84.5 & 89.2 & 1.4\\
MEGANet \cite{bui2024meganet}& 73.9 & 66.5 & 70.2 & 83.6 & 85.8 & 3.7 \\\hline
%ColnNet \cite{jain2023coinnet}  & 75.9 & 69.0 & 82.0 & 85.9 & 89.8 & 2.4 \\\hline
%SAM-mamba& 90.36 & 81.56 & 94.44 & 94.76 & 96.52 & 0.71  \\ \hline
%SAM-mamba-352& \textcolor{red}{90.21} & \textcolor{red}{80.99} & \textcolor{red}{94.03} & \textcolor{red}{95.40} & \textcolor{red}{98.21} & \textcolor{red}{0.68} \\ \hline
% SAM-mamba-352-self& \textcolor{red}{82.14} & \textcolor{red}{73.85} & \textcolor{red}{83.46}& \textcolor{red}{89.77}& \textcolor{red}{91.9}& \textcolor{red}{1.17} \\ \hline

%SAM-mamba-end-to-end& \textcolor{red}{82.2} & \textcolor{red}{73.9} & \textcolor{red}{83.5}& \textcolor{red}{89.8}& \textcolor{red}{91.9}& \textcolor{red}{1.2} \\ \hline
\textbf{SAM-Mamba} & \textcolor{red}{84.8} & \textcolor{red}{78.2} & \textcolor{red}{85.5}& \textcolor{red}{91.6}& \textcolor{red}{93.3}& \textcolor{red}{1.0} \\ 
\hline
\end{tabular}}
\end{table}
\vspace{-0.5cm}
 \begin{figure*}
  \centering
   \includegraphics[width=0.7\linewidth]{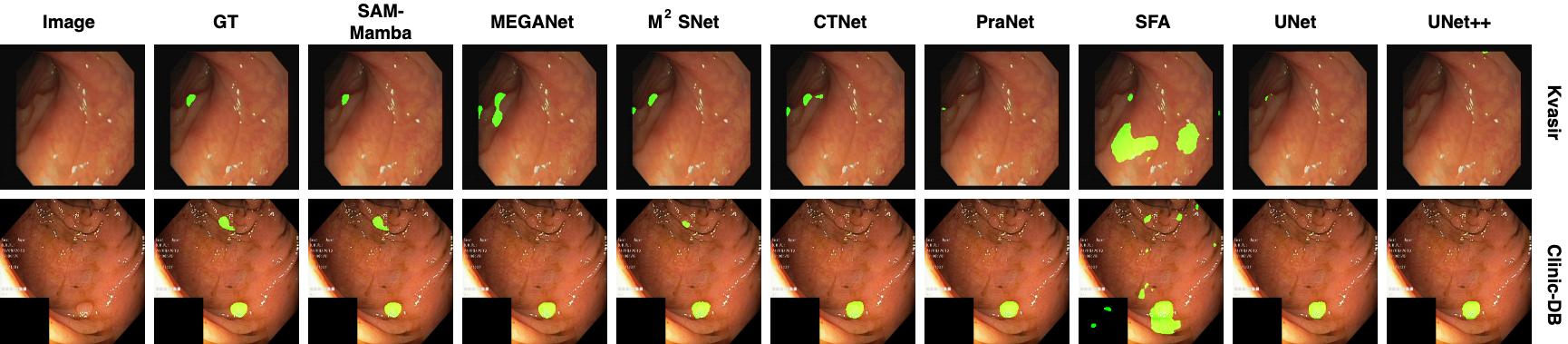}
   \caption{Qualitative comparison on seen datasets (Kvasir-SEG and CVC-ClinicDB), showcasing the model's ability to accurately segment polyps across diverse sizes, textures, and homogeneous regions.}
   \label{fig:seen}
\end{figure*}
%% \vspace{-0.5cm}
%% Table: CVC-ClinicDB (seen)
%
%%-------------------------------------------------------------------------
%
%%-------------------------------------------------------------------------
% \vspace{0.1cm}
\subsection{Qualitative Comparison}
\begin{figure*}[!h]
  \centering
   \includegraphics[width=0.68\linewidth]{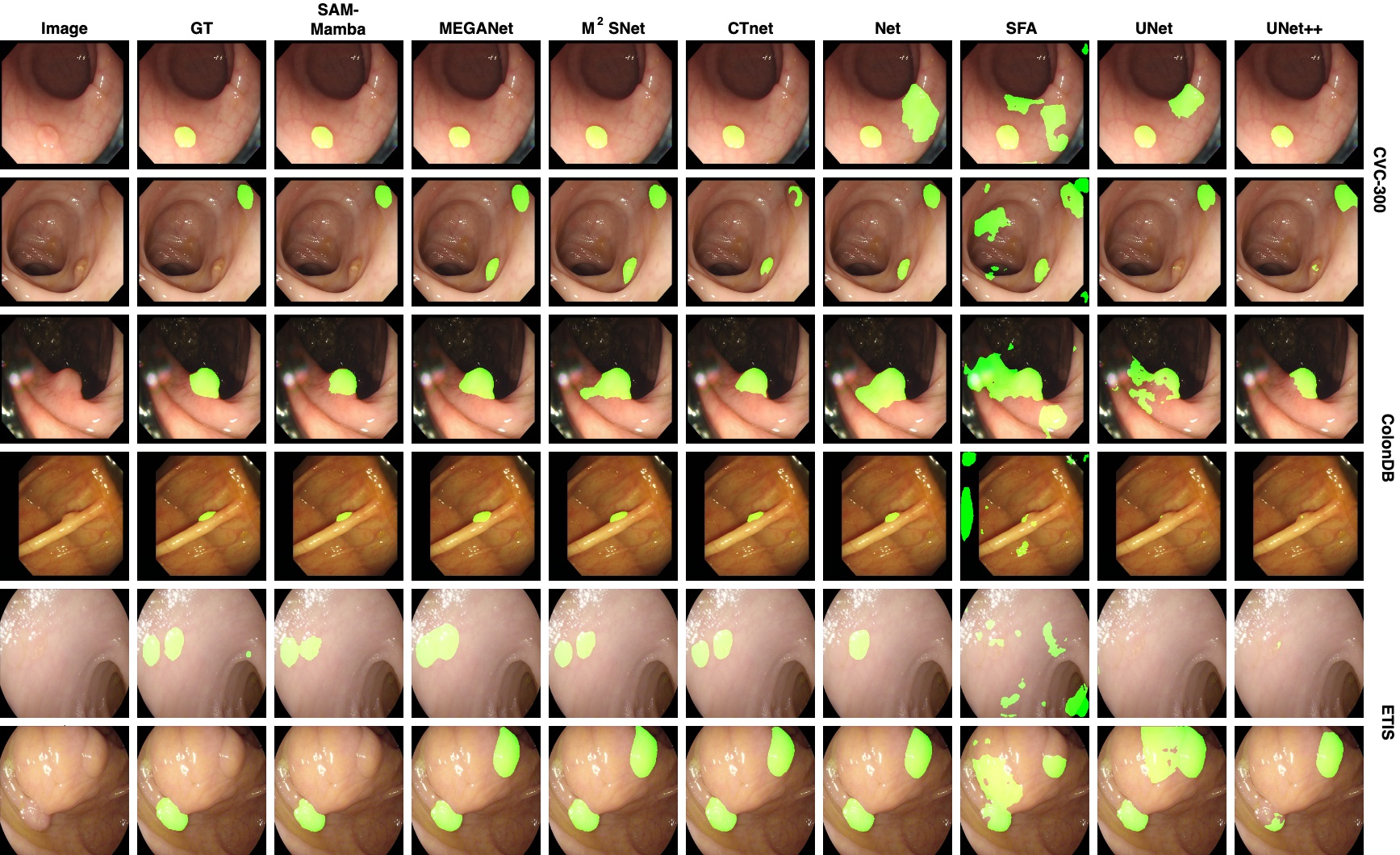}
   \caption{Qualitative comparison on unseen datasets (CVC-300, CVC-ColonDB, and ETIS), highlighting the model's superior generalization capabilities to accurately segment polyps of various sizes, textures, and homogeneous regions.}
   \label{fig:unseen}
\end{figure*}

\begin{figure*}
  \centering
   \includegraphics[width=0.65\linewidth]{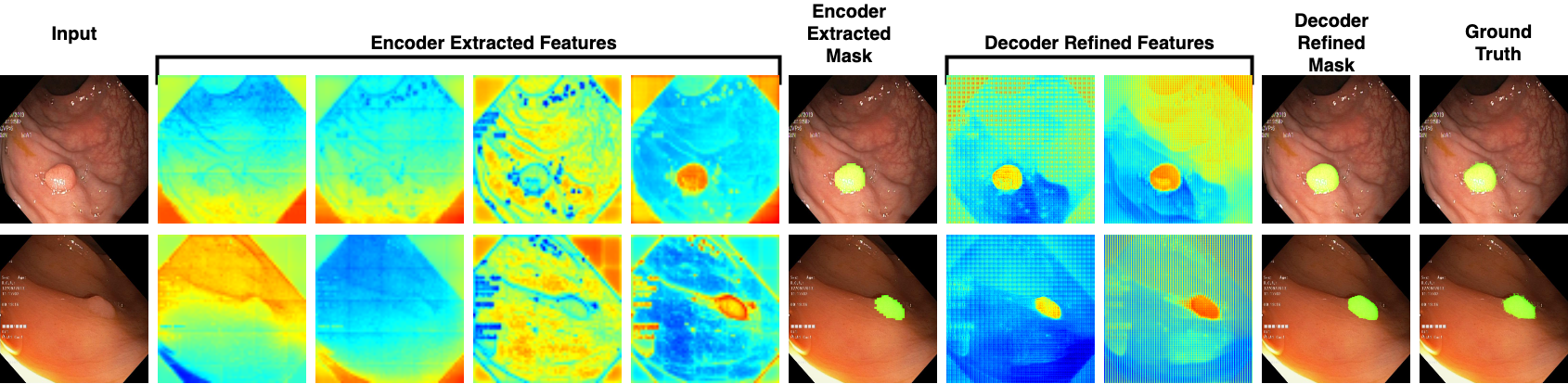}
   \caption{Illustration of the sequence learning progression within the SAM-Mamba model through a set of heatmap visualizations: the input image is followed by a set of encoder extracted features, the encoder's extracted mask, the decoder refined features, the refined segmentation mask, and the ground truth.
}
   \label{fig:heatmap}
% \vspace{-0.3cm}
\end{figure*}

Qualitative evaluations of our model are conducted on both seen and unseen datasets. As demonstrated in Figure \ref{fig:seen}, the MSD module enables our model to accurately identify the secondary polyp in the CVC-ClinicDB dataset. Similarly, on the Kvasir-SEG dataset, our model achieves a notably low false positive rate, attributed to the efficacy of the Mamba layer in exploring comprehensive global contextual information. For unseen datasets, such as CVC-300, our model consistently maintains a low false positive rate, paralleling the performance observed on seen datasets. This consistency is further illustrated in Figure \ref{fig:unseen} for the CVC-ColonDB and ETIS dataset, where our model demonstrates similar robustness. The learning progression in the encoder and decoder of our SAM-Mamba model is sequentially represented via a set of heatmap visualizations in Figure \ref{fig:heatmap}, demonstrating its robust learning capabilities.

In summary, the MSD module enhances our model's capability to detect polyps of varying sizes, while the Mamba-Prior module effectively reduces false positives, contributing to overall improved segmentation accuracy.

% \vspace{-0.4cm}
\subsection{Ablation Study and Discussion}
\paragraph{Effect of Mamba-Prior Components:} The ablation study results presented in Table \ref{table:1} highlight the impact of various components within the SAM-Mamba across different datasets. The results demonstrate a notable performance gain when incorporating the Mamba component into the SAM model. Specifically, the model with both MSD and Mamba adapters achieves superior results across all datasets, with mDice scores of 92.4\% on Kvasir-SEG, 94.2\% on CVC-ClinicDB, 85.3\% on CVC-ColonDB, 92.0\% on CVC-300, and 84.8\% on ETIS, while demanding additional 9.5\% parameters. In contrast, configurations without the Mamba component obtain a lower performance. This implies that the inclusion of the Mamba and adapter leads to a significant enhancement in the model's capability to capture more detailed and salient features, leading to improved segmentation results. The performance gained by Mamba corroborates its effectiveness on seen datasets and robustness towards unseen datasets.
% \vspace{-1cm}
\begin{table}[!h]
% \vspace{1cm}
    \caption{Results of ablation study on the effect of Mamba-Prior components. Here, D1:Kvasir-SEG, D2:CVC-CliniDB, D3:CVC-ColonDB, D4:CVC-300, D5:ETIS.}
    \centering
    \LARGE % Change the font size here
    \resizebox{\columnwidth}{!}{%
        \begin{tabular}{|c|c|c|c|c|c|c|c|c|}
            \hline
            \multirow{2}{*}{Adapter} & \multirow{2}{*}{MSD} & \multirow{2}{*}{Mamba} & \multirow{2}{*}{Params (M)} & \multicolumn{2}{c|}{Seen} & \multicolumn{3}{c|}{Unseen} \\
            \cline{5-9}
            & & & & D1 & D2 & D3 & D4 & D5 \\
            \hline
            \checkmark & - &- & 94 & 89.9 & 89.9 & 80.1 & 80.9 & 80.6 \\
            \checkmark & \checkmark &- & 101 & 90.9 & 91.3 & 80.8 & 90.3 & 81.2 \\
            \checkmark & \checkmark & \checkmark & 103 & 92.4 & 94.2 & 85.3 & 92.0 & 84.8 \\
            \hline
        \end{tabular}%
    }
    \label{table:1}
% \vspace{-0.6cm}
\end{table}
% \begin{table}[h!]
%     \centering
%     \resizebox{\columnwidth}{!}{\begin{tabular}{lccccc}
%         \toprule
%         \textbf{Model} & \textbf{Flops (G)} & \textbf{Parameters (M)} \\
%         \midrule
%         PVT-Cascade & 15 & 35 \\
%         CTNet & 45 & 25\\
%         Meganet & 44 & 23\\
%         SAM+Mamba & 4351 & 107\\
%         \bottomrule
%     \end{tabular}}
%     \caption{Ablation Study Results - Table 1}
%     \label{table:1}
% \end{table}
\vspace{-0.5cm}
\paragraph{Effect of kernel size:} In this experiment, we verify the effectiveness of different components of MSD. Table \ref{table:2} shows that the `Multi-scale +Mamba' configuration consistently outperforms uni-scale variants, especially on unseen datasets such as CVC-ColonDB, CVC-300, and ETIS. While uni-scale configurations perform well on seen datasets such as Kvasir-SEG (90.6) and CVC-ClinicDB (92.5), they struggle on unseen datasets (80.9 and 81.3), likely due to their fixed kernel size. The multi-scale approach improves generalization across different data scales, delivering the best results across all datasets and enhancing segmentation performance and robustness.
\begin{table}[!t]
    \caption{Results showing the effect of uni-scale \& multi-scale approaches within  MSD configuration. Here, D1: Kvasir-SEG, D2: CVC-ClinicDB, D3: CVC-ColonDB, D4: CVC-300, D5: ETIS.}
    \centering
    \resizebox{\columnwidth}{!}{%
    \begin{tabular}{|l|c|c|c|c|c|}
        \hline
        \multirow{2}{*}{Configuration} & \multicolumn{2}{c|}{Seen} & \multicolumn{3}{c|}{Unseen} \\
        \cline{2-6}
        &  D1 & D2 & D3 & D4 & D5 \\
        \hline
        Uni-scale $3\times 3$ +Mamba & 90.1 & 92.2 & 80.9 & 90.2 & 81.3 \\
        Uni-scale $5\times 5$ +Mamba & 90.6 & 92.5 & 80.9 & 90.1 & 80.9 \\
        Uni-scale $7\times 7$ +Mamba & 90.5 & 92.2 & 81.0 & 90.1 & 81.1 \\
        Multi-scale &  90.9 & 91.3 & 80.8 & 90.3 & 81.2 \\
        Multi-scale +Mamba & 92.4 & 94.2 & 85.3 & 92.0 & 84.8 \\
        \hline
    \end{tabular}}
    \label{table:2}
    \vspace{-0.5cm}
    
\end{table}
\section{Conclusion}
This paper presents a new method called \textbf{SAM-Mamba} for generalized zero-shot polyp segmentation. The primary innovation lies in integrating the Mamba-Prior module, which incorporates the Multi-scale Spatial Decomposition and dependency modeling of intra-scale features to extract polyps of varying shapes and sizes. Thanks to the increased ability of Mamba in modeling long-range feature dependencies, SAM-Mamba can effectively localize complex polyps and their boundaries in both seen and unseen datasets. Both quantitative and qualitative results on five benchmark datasets demonstrate the superior feature learning and generalization abilities of SAM-Mamba over traditional CNN, ViT, and Adapter-based models. 

%\vspace{-0.5cm}
\paragraph{Acknowledgement:} This work is supported by the Anusandhan National Research Foundation (ANRF), Department of Science and Technology, Government of India under project number CRG/2023/007397. D. Jha is supported by the University of South Dakota. 
%Experimental results on two seen and three unseen datasets during training highlight the efficacy of SAM-Mamba in polyp segmentation. , illustrates the benefits of the proposed SAM-Mamba.

%------------------------------------------------------------------------
% \section{Final copy}

% You must include your signed IEEE copyright release form when you submit your finished paper.
% We MUST have this form before your paper can be published in the proceedings.

% Please direct any questions to the production editor in charge of these proceedings at the IEEE Computer Society Press:
% \url{https://www.computer.org/about/contact}.

%%%%%%%%% REFERENCES
{\small
%\bibliographystyle{ieee_fullname}
%\bibliography{egbib}

}

\end{document}